\documentclass[letterpaper, 10 pt, conference]{ieeeconf}  

\IEEEoverridecommandlockouts                              

\usepackage{graphicx}
\usepackage{caption}
\usepackage{url}

\newcommand{\code}[1]{\textbf{\texttt{#1}}}

\title{\LARGE \bf
PyRIDE: An Interactive Development Environment for PR2 Robot
}

\author{Xun Wang$^{1}$, Mary-Anne Williams$^{1}$
\thanks{$^{1}$Xun Wang and Mary-Anne Williams are from Social Robotics Studio, Centre of Quantum Computation and Intelligent Systems, University of Technology, Sydney, Australia
        {\tt\small xun.wang-1@uts.edu.au}}%
}

\begin{document}

\maketitle
\thispagestyle{empty}
\pagestyle{empty}

\begin{abstract}


Python based Robot Interactive Development Environment (PyRIDE) is a software that supports rapid \textit{interactive} programming of robot skills and behaviours on PR2/ROS (Robot Operating System) platform. One of the key features of PyRIDE is its interactive remotely accessible Python console that allows its users to program robots \textit{online} and in \textit{realtime} in the same way as using the standard Python interactive interpreter. It allows programs to be modified while they are running. PyRIDE is also a software integration framework that abstracts and aggregates disparate low level ROS software modules, e.g. arm joint motor controllers, and exposes their functionalities through a unified Python programming interface. PR2 programmers are able to experiment and develop robot behaviours without dealing with specific  details of accessing underlying softwares and hardwares. PyRIDE provides a client-server mechanism that allows remote user access of the robot functionalities, e.g. remote robot monitoring and control, access real-time robot camera image data etc. This enables multi-modal human robot interactions using different devices and user interfaces. All these features are seamlessly integrated into one lightweight and portable middleware package. In this paper, we use four real life scenarios to demonstrate PyRIDE key features and illustrate the usefulness of software.
\end{abstract}

\section{INTRODUCTION}
Over the past decades, robotic technologies and robot systems have been steadily becoming more sophisticated and complex. Programming of an advanced robot system such as the PR2 robot remains a challenging task. There are three long standing issues that still inhibit rapid advancement in robot software development. First, programming for robots usually requires specialist knowledge of the robot platform involved. Robot programmers need to have in-depth understanding of both robot hardware and software. Programmers must have detailed knowledge of the existing software modules written specifically for the platform. In some cases, programmers may even have to learn the specialised programming language used for the robot platform\cite{Baillie2005}. These requirements introduce a steep learning curve for novice programmers and create barriers for expert robotic programmers to transfer their expertises and software to different robot platforms. Second, highly customised software development environments are required before the start of robotic software development. Setting up such a development environment is both technically challenging and time consuming for unexperienced users. The combination of these issues and the difficulty of gaining access to physical robots deters the entry of novice robotic programmers and constraints the growth of the robot software development community. Finally, integration of disparate software modules to produce practically useful robot functions is also nontrivial and time consuming.

The recent development and wide adoption of Robot Operating System (ROS)\cite{Quigley2009} has partly elevated first and third issue by ``standardising" the software module packaging and providing a common data communication framework between different modules. Software implementation of common robotic functions and algorithms can now be shared among different robot platforms. ROS uses C++ and Python as the main programming languages and provides a set of APIs for both languages so that programmers with background of these languages can adopt ROS quickly. The problem, however, is that programming in ROS is still a non-trivial exercise and setting up the ROS development environment (on a system other than Ubuntu Linux) can be both challenging and time consuming. On the other hand, there are development environments \cite{Aldebaran,Jackson2007,Resnick2009} that are easy to setup and use visually based programming techniques to ease the difficulties of programming robot routines. However, these systems are often limited to certain platforms and far less flexible than traditional programming approaches.

In this paper, we introduce a lightweight middleware, Python based Robot \textit{Interactive} Development Environment (PyRIDE), that can considerably ease the burdens of programming advanced robots, specifically the PR2 robot. PyRIDE is designed to address the previously discussed issues in small and practically useful steps: a) provide an abstraction layer to integrate the existing robot software functionalities and expose them through a unified Python robotic programming interface. It is no longer necessary for novice PR2 programmers to learn specific details of ROS and how to communicate with low level PR2 software modules before they can start developing complicate robot skills and behaviours on PR2. Expert PR2/ROS developers can also easily integrate existing ROS modules with PyRIDE through this abstraction layer.
b) provide an interactive shell facility that allows programmers to experiment and test code on robots remotely and interactively in realtime. This also makes client side development software installation redundant. The overhead of accessing and administering a PR2 robot is minimised because users no longer require local user accounts on the robot's computers before they can install and execute their code on the robot\footnote{We assume that the robot resides within an isolated and secure network environment for its software development.}, and c) provide a client-sever mechanism so that programmers can develop remote client programs quickly without dealing with tasks such as data communication between the client and robot.

In the following sections, we will first present the system architecture of PyRIDE and discuss the key design choices we made for the system. We will then use four practical use cases to demonstrate the effectiveness of the system. Finally, we conclude with a brief discussion on the limitations and possible future improvements for PyRIDE. We would like to emphasise that PyRIDE is not limited to PR2/ROS. PyRIDE also operates on NAO robots. It can be easily ported to other robot platforms.

\section{SYSTEM ARCHITECTURE}
\begin{figure}
\centering
\includegraphics[width=1\linewidth]{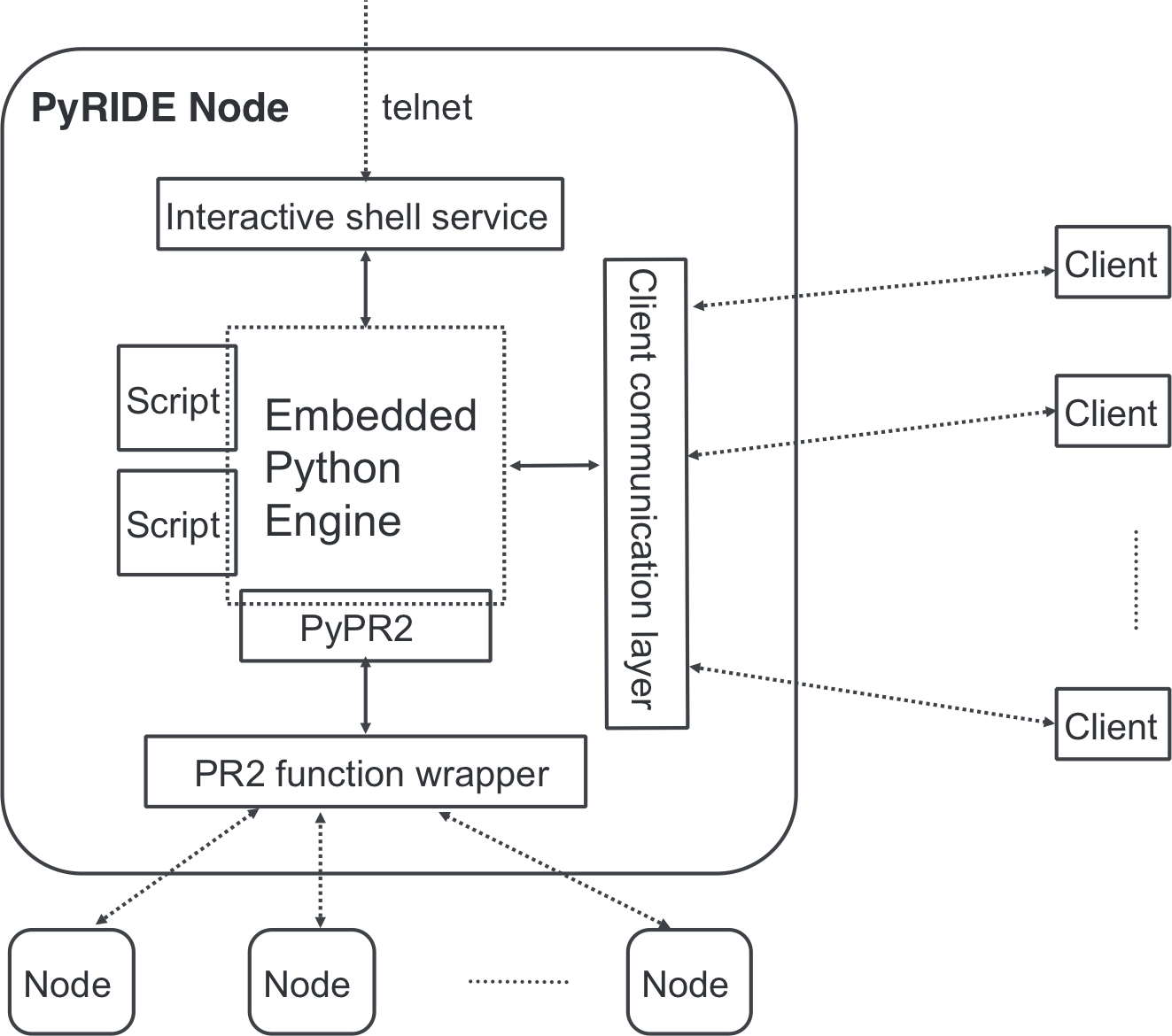}
\caption{A System architecture diagram for PyRIDE on ROS/PR2. Note: dashed arrow lines are network communication.}
\label{sys_arch}
\end{figure}

The system architecture of PyRIDE is shown in Fig. \ref{sys_arch}. PyRIDE can be viewed as a lightweight middleware that sits on the top of an existing robot software environment. It provides an abstraction layer for the existing software modules and unify their functionalities into a common Python programming interface. The PyRIDE software architecture is agnostic to the underlying robot platform. PyRIDE can be packaged in accordance to the underlying platform requirements. Fig. \ref{sys_arch} shows that PyRIDE for PR2 works under the ROS environment as a standard ROS node. On the NAO robot, PyRIDE is packaged as a loadable runtime NAOQi module that automatically connects to other runtime modules when it is loaded. Since the core of PyRIDE is written in standard portable C++ and C, it can be ported to other robot platforms with relative ease.

\subsection{System Components}
\begin{figure}
\centering
\includegraphics[width=0.95\linewidth]{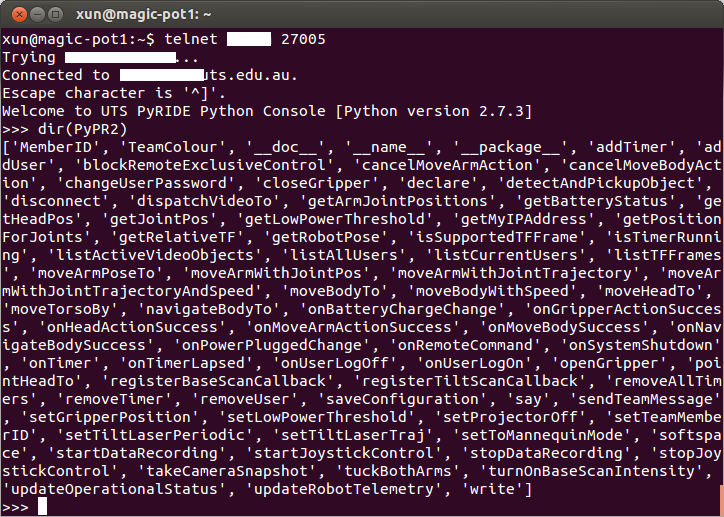}
\caption{A telnet session of the PyRIDE interactive interface. It shows the array of PR2 specific functions provided through \textbf{PyPR2} extension module.}
\label{telnet_console}
\end{figure}

\begin{figure}
\centering
\includegraphics[width=0.95\linewidth]{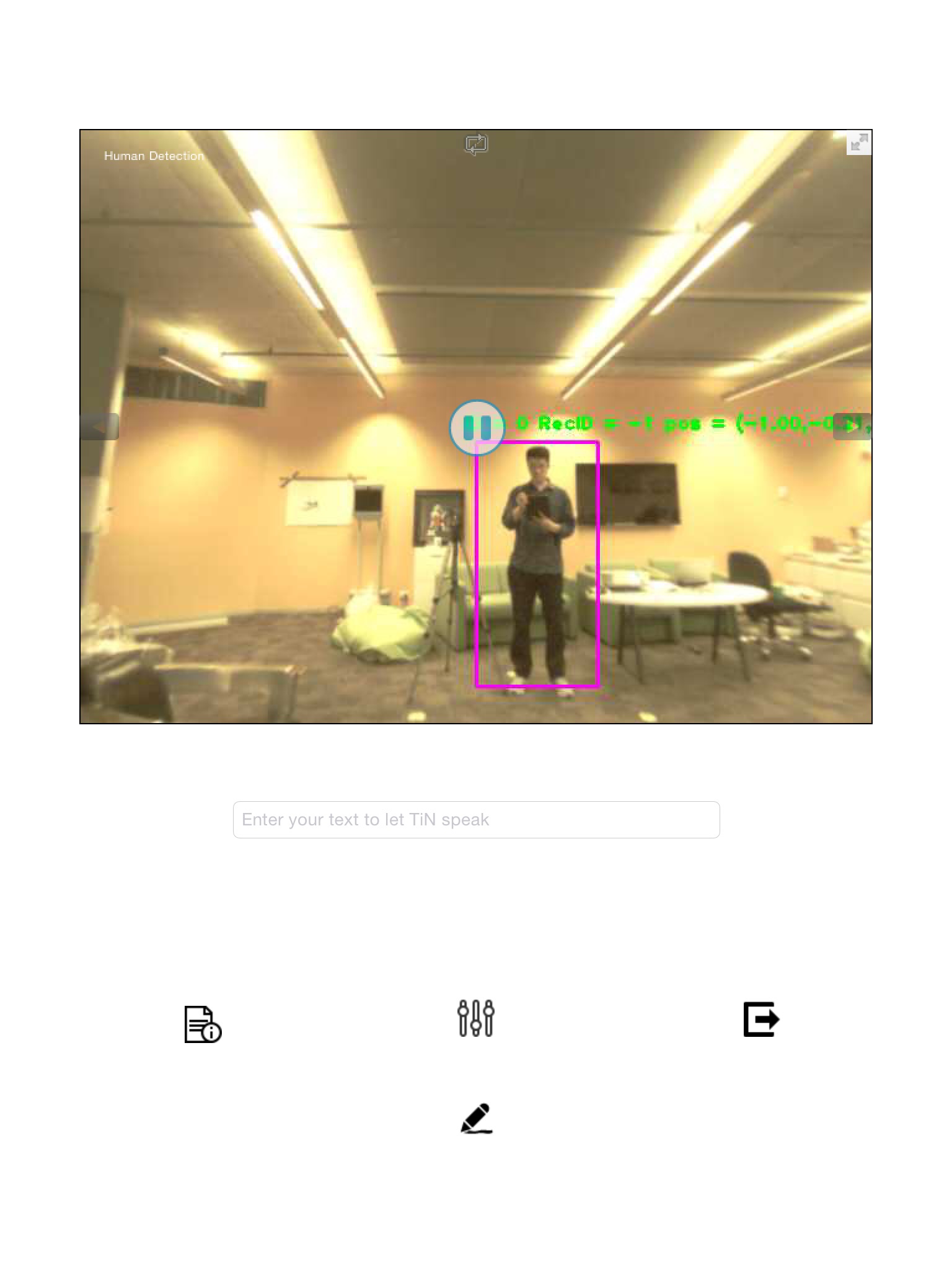}
\caption{A live screenshot of a PyRIDE remote iPad client. The screenshot shows a testing session for a human detection subsystem running on our PR2 robot.}
\label{ipad_client}
\end{figure}

At the centre of PyRIDE is an embedded Python scripting engine that provides a self contained programming environment for its users. Functional access to the robot system is provided through a single \textbf{PyPR2} Python module interface. PR2 programmers can interact with the robot system by calling appropriate methods in \textbf{PyPR2} and setting up necessary callback functions to receive information from the rest of system and other third-party ROS nodes. There are five key system components attached to the embedded Python engine:

\begin{itemize}
\item \textbf{Interactive shell interface} that provides remote programming access to the embedded Python engine. The interface supports basic telnet network communication protocol so that a programmer can interact with the Python engine remotely using a simple telnet client. This facility provides the same interactivity as a standard interactive Python interpreter (Fig. \ref{telnet_console}).
\\
\item \textbf{Robot function wrapper and Python extension module}. This is the only system component in PyRIDE that is tailored to the specifics of a robot platform. The robot function wrapper such as PR2 function wrapper manages the lower level communication with the PR2 specific subsystems, e.g. joint\_state\_controller. It provides the high level wrapper functions to the underlying functionalities. These wrapper functions are further exposed to the embedded Python engine through \textbf{PyPR2} extension module (see Fig. \ref{telnet_console}). For example, one can call \code{PyPR2.moveArmWithJointPos} method in the remote interactive shell to move a PR2 arm joints to specific joint angles. PyRIDE defines an abstract programming interface so that this platform specific functional wrapper module can be tailored to a different robot platforms.
\\
\item \textbf{Client communication layer}. PyRIDE uses the standard client server model to provide remote user access to the robot platform. This communication subsystem enables data exchanges between the robot and remote client programs, for example, real-time video and audio streaming to a remote client.
All remote client communications are managed through the scripts running on the embedded Python engine.
\\
\item \textbf{Scripts}  loaded and executed in the embedded Python engine. These custom Python scripts are developed by programmers for advanced robot behaviours using the services provided by PyRIDE. Each script (or a set of functionally related scripts) can be considered as a small application that runs on the top of PyRIDE middleware. These applications can be loaded and executed dynamically within PyRIDE. Note that, PyRIDE automatically bootstraps a Python startup script \code{py\_main.py}.
\\
\item \textbf{Clients} are the remote user programs for PyRIDE. Remote users can use the data exchange functionalities provided by the client to communicate with the PyRIDE service running on a robot platform. The PyRIDE client component supports multiple operating systems and it offers a set of programming interface so that programmers can customise and extend the clients accordingly. Fig. \ref{ipad_client} shows an example of PyRIDE remote client running on an iPad device. Camera images are streaming to the client in realtime.

\end{itemize}

\subsection{Design Choices}
The design and development of PyRIDE represents a small important step to further improve the programming environments for the PR2 (and other) robots. We attempted to balance simplicity and easy to use with the need of programming flexibility and support for extensive testing. A number of deliberate design decisions were made to achieve our goals.

First, similar to ROS and other systems\cite{Blank2004}, we adopt Python as the main programming language due to its clear syntax and is easy to learn and use. There is already a large population of Python developers. Unlike the existing systems such as ROS that incorporate Python through the standard module extension mechanism, PyRIDE embeds a Python interpreter directly at its core and it becomes a self-contained development environment. Combined with a remote interactive shell service that uses the standard telnet communication protocol\cite{Postel1983}, PyRIDE provides the same interactive programming facility as the standard interactive Python interpreter (Fig. \ref{telnet_console}). This means that a programmer has direct remote access to the functionalities provided by PyRIDE without installing additional software on his/her local development machine, apart from telnet. No user specific account needs to be created and maintained on the PR2 robot's computer in order for a new user to start programming on the robot. It also means that one can develop programs on PyRIDE from (almost any) personal computer with a network connection to the robot\footnote{The interactive shell service can be switched off so to prevent unauthorised access after software development moves to the deployment phase.}. PyRIDE helps a PR2 programmer to focus on his/her work by minimising communication and administrative overhead and removing distractions such as working with the underlying Linux system.

Second, the integration framework with robot function wrapper brings the existing robot functions together under one unified programming interface so that programmers can immediately utilise them without any additional efforts to set up correct access. For example, to obtain the current joint positions of a PR2 robot arm, one can simply call \code{PyPR2.getArmJointPositions} method in PyRIDE Python environment. In contrast, the standard approach in ROS requires many lines of boilerplate code to set up necessary communication proxy channel to the appropriate ROS node and run an event loop in order to retrieve the relevant data. The standard ROS programming approach becomes cumbersome and error prone when a programmer starts to develop complex robot behaviours that involve the coordination and execution of multiple underlying ROS nodes. With the PyRIDE function integration framework, advanced PR2 developers can integrate their existing software modules together under PyRIDE and expose their functionalities to a significantly broader programmer audience.

Third, PyRIDE is designed to automatically search and load a default Python script on startup. Through this  bootstrap mechanism, any scripts developed in the embedded Python environment can be immediately deployed. PyRIDE enables seamless transition between software development and software deployment. A typical software development to deployment in PyRIDE starts with the development and testing of code using the interactive mode. Afterwards, the functioning code is packaged into modules and linked to the startup script. PyRIDE will automatically load and execute the code on start up. There is no complicate deployment procedure.

Finally, we developed a client server subsystem in PyRIDE to enable human robot interactions with multiple remote devices and user interfaces. Remote client component, e.g. \code{pyride\_remote}, can be used to develop custom remote client applications to control and monitor robot operation in realtime. Such remote access is integrated and managed directly through scripts running in PyRIDE. This subsystem enhances the accessibility of the robot considerably from both user and programmer perspective.

In summary, we have taken the aforementioned steps to make programming a PR2 robot significantly easier. Each individual features and ideas that we have adopted in PyRIDE is not completely new by itself. Our innovations focused on the seamless integration of these features and ideas to provide practical improvements in terms of robot accessibility, ease of use and easy integration of disparate software components. In the next section, we will use four real life use cases to show that the developer focused improvements in PyRIDE increase the efficiency of robot experimentation, programming and testing. We will show PyRIDE is robust and flexible enough to have highly complex robot software built upon it.

\section{USE CASES}
In this section, we present four real life scenarios to demonstrate the capabilities of PyRIDE described in the previous sections. We use the PR2 robot in our research lab in these examples. Each scenario is shown in the video clips accompanied with this paper. The video is available at \url{http://youtu.be/0DTB62lm8z4}.

\subsection{Experimental Setup and Data Collection}
One of the common tasks in robotic research is setting up an experimental environment for a robot to collect various sensor data, e.g. video stream, for offline analysis and modeling. This experiment preparation and data collection process is usually composed of disparate arrays of operating procedures that are often put together in an `ad-hoc' fashion. For instance, suppose we would like to manually move the PR2 robot to a location, then make the robot arm perform some pre-defined actions, while the robot sensor data is being recorded. To manipulate the robot using the standard approach involves manual execution of various command line tools and custom made scripts under multiple terminal consoles in a specific pre-planned sequence. Minor adjustments in the experiment setup will quickly disrupt the pre-planned experimental sequence, and making small changes to the plan, e.g. modifying action sequence, is both time consuming and error prone.

This experimental setup procedure can be performed in a straight forward fashion with PyRIDE through its interactive Python console:
\begin{enumerate}
\item \code{telnet pr2 27005}.\\Connect to the interactive shell service remotely.
\item \code{PyPR2.tuckBothArms()}.\\Fold the arms and ready for moving the robot.
\item \code{PyPR2.startJoystickControl()}.\\Enable PS3 joystick control of PR2 and move the robot to the experimental location.
\item \code{PyPR2.stopJoystickControl()}.\\Disable the joystick control of PR2.
\item \code{PyPR2.moveTorsoBy(0.02)}.\\Adjust the height of the robot.
\item \code{PyPR2.moveHeadTo(`base\_link',5.0,0.0, 1.2)}.\\Adjust the head position.
\item \code{PyPR2.setToMannequinMode(True)}.\\Set to mannequin mode so that robot arms can be move manually to a pre-defined position.
\item \code{PyPR2.setToMannequinMode(False)}.\\Turn off mannequin mode and restore normal arm control.
\item \code{PyPR2.setTiltLaserPeriodic(0.5,2.0)}.\\Activate the tilt laser scanner.
\item \code{PyPR2.startDataRecording(REC\_CAM| REC\_SCAN|REC\_TF)}.\\Start recording sensory data.
\item Perform some predefined actions.
\item \code{PyPR2.stopDataRecording()}.\\Stop the data recording.
\end{enumerate}

All steps above are demonstrated live in the video clips 1 and 2. It is clear that the interactive programmable control of various robot functions has removed many nuances in operating the PR2 robot. This significantly improves the management of the robot and streamlines experimental setup and data collection processes.

\subsection{An Inverse Kinematics System}
The previous example shows some of the basic functionalities offered by PyRIDE. Highly sophisticated software can be built entirely within PyRIDE. In fact, it has been used to build a new optimised inverse kinematic control system, \textbf{S-PR2}, for the PR2 robot\cite{Taghiabadi2015}. This system is written entirely in Python\footnote{S-PR2 uses scientific computing Python extension \code{numpy} that is written largely in C.}. It uses the joint redundancies to generate optimised smooth joint trajectories in 2D and 3D world spaces. PyRIDE provides the realtime joint information to the S-PR2 system and takes computed low-level joint control commands from S-PR2 to drive the robot arms. Combined with a \textit{PyRIDE remote iPad client}, we were able to develop an integrated system that allows the PR2 to copy human handwriting from the iPad client inputs to a whiteboard. The third video clip shows a demonstration of this handwriting system.

\subsection{Realtime Code Debugging}
Diagnosing and testing robot functions is one of the most challenging tasks in programming robots. Due to the realtime nature of robotic systems, it is difficult to pause and diagnose code related problems while the robot programs are running continuously. A crash or a restart of a subsystem often means that the entire (or a large part of) robot system needs to be reinitialised. PyRIDE offers an alternative approach to the traditional code and test software development cycle. First, PyRIDE is fault tolerate since crashes of individual scripts will not bring down the entire system. All debugging and crash logs are saved in a log file through a realtime logging subsystem in the PyRIDE. The interactive nature of PyRIDE encourages agile code development so that individual functions are coded and tested separately in rapid iterations. With a proper modular design, it is possible to execute, (manually) pause, fix code bugs, (manually) reload the portion of code that contains the fixes and finally continue the execution without completely restarting the application and losing vital system state information. This iterative code and test approach was adopted in the development of S-PR2. Consequently, the development productivity was dramatically improved.

The PyRIDE remote clients also provide additional facilities to support realtime diagnosis of code problems. In particular, a remote client on a mobile device can greatly increase the mobility of programmers when they conduct field tests. Custom robot operational data can be streamed to the mobile device, the programmer is no longer bonded to a computer console that is often difficult to carry during testing. For example, when we test a human detection algorithm on the PR2 robot, the detection results are streamed to an iPad remote client (Fig. \ref{ipad_client}), while we are in front of the robot as the testing subjects (see video clip 4). It would be awkward to use a standard laptop computer in these situations.

\subsection{Third-party ROS Module Integration}
As an easy to use middleware, PyRIDE is not ideal platform for advanced programming tasks that require intensive computational resource or very low level realtime motor control. These tasks should be implemented in native C++ code for computational efficiency. However, one key use of PyRIDE is system integration. That is, advanced PR2/ROS developers can use PyRIDE as an integration tool for the existing ROS modules. Integration of a third-party ROS module with PyRIDE on PR2 involves modifications of PyRIDE PR2 function wrapper (\code{PR2ProxyManager.cpp}\footnote{See PyRIDE on PR2/ROS source code for details}) and the associated Python extension module (\code{PyPR2Module.cpp}). One can follow the standard ROS node initialisation procedure to set up the proxy to the third-party ROS node, e.g. PointHeadAction client, or subscribe to the topics provided by the node under \code{PR2ProxyManager::initWithNodeHandle} method. Public visible methods should be added to \code{PR2ProxyManager} to access the services provided by the ROS node. For example, \code{PR2ProxyManager::pointHeadTo} method uses the PointHeadAction client to control PR2 head direction. These publicly accessible methods can then be wrapped under a Python C extension function in \code{PyPR2Module.cpp}  to be accessed from the embedded Python engine in PyRIDE. See \code{PyModule\_PR2PointHeadTo} for example.

For ROS node data subscription, all data messages from a subscribed ROS node publisher are processed through callback functions and passed on to Python via  \code{PyPR2Module::invokeCallback} function after  data transformation. See \code{PR2ProxyManager::tiltScanDataCB} for example. A simpler alternative for sending data to PyRIDE is publishing the data to the topic of \code{pyride\_pr2/node\_status} as formatted strings. PyRIDE continuously listens on this topic and pass received messages to a Python callback function linked to \code{PyPR2.onNodeStatusUpdate}. Messages can be then decoded in the callback function for further processing. In short, integrating a third-party ROS node with PyRIDE is very much a straight forward process.

\section{DISCUSSION}
We have presented a software development middleware for programming high-level robot skills and behaviours on PR2 robot. The design philosophy of PyRIDE is to take small and practical steps to improve the software development process for the robot. Key features such as the remote interactive Python shell service allow a PR2 programmer to operate, program and perform experiments on PR2 with significantly greater efficiency compared with the standard ROS programming approach. PyRIDE lowers the entry level for robot programming by minimising the steep robot programming learning curve and avoiding cumbersome development environment setup. A Python programmer can directly apply his/her existing skills to program a PR2 without dealing with the specific details of ROS and the underlying operating system. In comparison, a combination of rosh shell\cite{rosh2010} and IPython\cite{Perez2007} offer an interactive programming experience similar to PyRIDE. However, one still needs to install the necessary software on his/her local development machine or have access to a well maintained user account on the robot. One still requires in-depth knowledge of ROS and the PR2 specific ROS modules, e.g. what plugin should be used, what ROS service/message type should be used, before he/she can begin his/her work on PR2. PyRIDE helps PR2 programmers to concentrate on the programming tasks at hands by removing all these unnecessary ``distractions".

PyRIDE works as a smart integration tool for combining the functionalities provided by various ROS modules. A third-party ROS node integrated with PyRIDE can be dynamically managed and utilised from the embedded Python engine with minimum setup. Sophisticate robot behaviours can be developed using relatively simple scripts that aggregate various services provided by the external ROS nodes. Third-party integration with PyRIDE is no more complicated than the standard ROS programming. PyRIDE could be a very useful tool even for the most advanced PR2 developers.

The modular design of PyRIDE means it is agnostic to the underlying robot platform. In addition to PR2/ROS, PyRIDE supports NAO humanoid robot. In fact, PyRIDE was used to develop our NAO robot  soccer system that competed in RoboCup 2011\cite{Stanton2011}. PyRIDE can be ported to other robot systems similar to PR2 and NAO with minimal efforts. Furthermore, with the popularity of mobile devices and associated robot toys like Romo\cite{Romotive}, we have ported PyRIDE onto the iOS/Romo platform\footnote{We plan to release PyRIDE on Romo as an app in iOS App Store in near future.} so that even robot hobbyists and school children can develop programs for their devices. PyRIDE on Romo/IOS provides the same coding environment as PyRIDE on PR2/ROS and PyRIDE on NAO. Script modules that are not specifically tied to the underlying platform can be directly shared among the systems running PyRIDE.  The common programming approach under PyRIDE means even code that uses platform specific functions can be adapted to different platforms without lengthy rework. In another words, PyRIDE establishes accessible pathway for robot programs and robot software developers to migrate between different platforms. It is our hope that PyRIDE can help to attract more developers to robot software development because it significantly reduces the usual high-bar of expertise needed and make robotics more accessible to a much larger audience of software engineers and programmers.

It is important to highlight that features and ideas embodied within PyRIDE maybe be found in other systems. The innovation of PyRIDE is the seamless and intelligent combination of these features to improve programming efficiency and provide a much better programming experiences for both novice and advanced PR2 developers. Like any software systems, there are several areas where PyRIDE can be further improved. First, PyRIDE only provides a minimal user interface. This could be considered as a drawback for the developers who are comfortable or prefer with Graphical User Interfaces. We will integrate PyRIDE with popular IPython/Jupyter\cite{Ragan-Kelley2014} web interface and investigate possible integration with development environments such as Eclipse. Second, even though PyRIDE has provided a software integration framework for robot developers, the task of integrating disparate software module requires some work. We plan to develop a software integration plugin system and working examples for developers who wish to integrate their systems with PyRIDE. Similarly, the PyRIDE remote client programming interface needs to be further consolidated for third party developers to write their own remote client tools upon PyRIDE remote client component. To this end, we are in the process of documenting and releasing parts of PyRIDE to the public domain so that people from the robot development community can help us improving the current software. In the current initial release phase, we have made PyRIDE on PR2/ROS source code available at \url{http://github.com/uts-magic-lab/pyride_pr2}, PyRIDE on NAO at \url{http://github.com/uts-magic-lab/pyride_nao}. We invite feedbacks and suggestions from the community.

\addtolength{\textheight}{-12cm}   

\bibliographystyle{IEEEtran} 
\bibliography{pyride}

\end{document}